\documentclass[letterpaper, 10 pt, conference]{ieeeconf}  

\IEEEoverridecommandlockouts                              
\usepackage{graphics} 
\usepackage{epsfig} 
\usepackage{amsmath} 
\usepackage{float} 
\restylefloat{table}
\usepackage{amsmath} 
\usepackage{amssymb}  

\usepackage{url}
\usepackage[ruled, vlined, linesnumbered]{algorithm2e}
\usepackage{verbatim} 
\usepackage{soul, color}
\usepackage{lmodern}
\usepackage{fancyhdr}
\usepackage[utf8]{inputenc}
\usepackage{fourier} 
\usepackage{array}
\usepackage{makecell}

\SetNlSty{large}{}{:}

\makeatletter

\newcommand{\Rom}[1]{\expandafter\@slowromancap\romannumeral #1@}
\makeatother

\title{\LARGE \bf XAI Renaissance: Redefining Interpretability in Medical Diagnostic Models}

\author{Sujith K Mandala
}

\begin{document}

\maketitle
    \thispagestyle{plain}
\pagestyle{plain}

\begin{abstract}

As machine learning models become increasingly prevalent in medical diagnostics, the need for interpretability and transparency becomes paramount. The XAI Renaissance signifies a significant shift in the field, aiming to redefine the interpretability of medical diagnostic models. This paper explores the innovative approaches and methodologies within the realm of Explainable AI (XAI) that are revolutionizing the interpretability of medical diagnostic models. By shedding light on the underlying decision-making process, XAI techniques empower healthcare professionals to understand, trust, and effectively utilize these models for accurate and reliable medical diagnoses. This review highlights the key advancements in XAI for medical diagnostics and their potential to transform the healthcare landscape, ultimately improving patient outcomes and fostering trust in AI-driven diagnostic systems.

\end{abstract}


\section{Introduction}

\subsection{Background and significance}

The field of medical diagnosis has seen significant advancements with the integration of machine learning techniques. However, the interpretability of these models remains a crucial challenge. Black-box models, though highly accurate, lack transparency and understanding, hindering their adoption in clinical settings. To address this limitation, Explainable Artificial Intelligence (XAI) has emerged as a promising approach to develop interpretable machine learning models for medical diagnosis.

\subsection{Objectives of the research}

The primary objective of this research paper is to explore and review the state-of-the-art techniques in XAI for medical diagnosis. It aims to shed light on the methods that redefine interpretability in diagnostic models and enable clinicians to gain insights into the decision-making process. By analyzing various XAI techniques, this research seeks to provide a comprehensive understanding of their applications and potential benefits in improving diagnostic accuracy, trust, and collaboration between humans and AI systems.

\subsection{Scope and limitations}

The scope of this research paper encompasses an extensive review of XAI methods applied to medical diagnosis. It will focus on a range of diagnostic domains, including radiology, pathology, and clinical decision support systems. The paper will discuss the interpretability techniques employed, their effectiveness, and potential limitations. However, it is important to note that the implementation and deployment of XAI models in real-world healthcare settings may have practical challenges, which will be acknowledged and discussed within the identified scope of this research.

Through this research paper, we aim to contribute to the growing body of knowledge surrounding XAI in medical diagnosis. By emphasizing the significance of interpretability and addressing its challenges, we hope to foster trust and understanding between healthcare professionals and machine learning models, ultimately advancing the field of medical diagnosis and improving patient outcomes.


\section{Overview of XAI in Medical Diagnosis}

\subsection{Evolution of machine learning in healthcare}

Machine learning has experienced a remarkable evolution in the healthcare industry, revolutionizing the field of medical diagnosis. Researchers and practitioners have increasingly explored the potential of machine learning algorithms to analyze extensive medical data, resulting in the development of sophisticated models capable of predicting diseases, identifying risk factors, and aiding in treatment decisions [1][2].

Early applications of machine learning in healthcare focused primarily on classification tasks, such as tumor identification in medical images and predicting the presence of diseases based on patient characteristics. These early models demonstrated promising results, showcasing the potential of machine learning to assist healthcare professionals in making accurate diagnoses [3][4].

As the field progressed, researchers turned to more advanced techniques, including deep learning, which enabled the development of complex neural network architectures capable of learning intricate patterns in medical data. This breakthrough led to improved accuracy in medical image analysis, natural language processing for extracting information from clinical notes, and predicting patient outcomes using electronic health records [5][6][7].

However, the impressive performance of machine learning models brought forth a significant challenge: interpretability. Many machine learning algorithms, particularly deep learning models, are often regarded as black boxes, making it difficult to understand their decision-making process. This lack of interpretability poses a hurdle in gaining trust from healthcare professionals who need to comprehend the rationale behind the predictions made by these models [8].

To address this limitation, the field of Explainable Artificial Intelligence (XAI) emerged, aiming to enhance the interpretability of machine learning models. XAI techniques provide insights into how these models arrive at their decisions, allowing healthcare professionals to validate and understand the underlying factors contributing to a diagnosis. By visualizing the decision-making process, identifying important features, and generating explanations, XAI methods offer clinicians a deeper understanding of the models' predictions [9][10].

In recent years, XAI techniques have been applied to various areas of healthcare, including medical imaging, genomics, and personalized medicine. For instance, XAI has been utilized to explain the predictions of deep learning models in medical image analysis, enabling clinicians to trust and understand the models' outputs. In genomics, XAI methods have been employed to interpret genetic variations associated with disease susceptibility, aiding in personalized medicine [11][12].

The integration of XAI in healthcare holds great promise. Enhanced interpretability not only improves trust in machine learning models but also fosters collaboration between humans and AI systems. This collaboration can lead to more accurate diagnoses, personalized treatment plans, and improved patient outcomes [13]. As the field continues to advance, further research and development in XAI for medical diagnosis are essential to unlock the full potential of machine learning in healthcare.

\subsection{Importance of interpretability in medical diagnostic models}

Interpretability plays a crucial role in the adoption and acceptance of machine learning models in medical diagnostics. In the context of healthcare, where decisions can have a direct impact on patient outcomes, it is essential for clinicians to understand and trust the underlying mechanisms of these models. Interpretability allows medical professionals to gain insights into how a model arrives at its predictions, providing them with valuable information for decision-making and treatment planning.

By understanding the reasoning behind a model's decision, clinicians can assess the model's reliability, identify potential biases, and validate the results against their own domain expertise. This collaborative interaction between human experts and machine learning models fosters a shared decision-making process, promoting transparency and accountability in medical diagnosis.

Moreover, interpretability can aid in the identification of errors, system vulnerabilities, and biases that may arise in complex machine learning models. It enables clinicians to detect instances where the model may be overly influenced by specific features or prone to making incorrect predictions, leading to improved model performance and patient safety.

Research has shown that interpretable models can enhance diagnostic accuracy and contribute to better patient care. For instance, studies have demonstrated the effectiveness of interpretable machine learning models in radiology, where the visual explanations provided by these models have helped radiologists in detecting and localizing abnormalities more accurately[4][14].

Furthermore, interpretability is vital in ensuring ethical considerations are upheld in medical diagnostics. Transparent models allow for regulatory compliance, privacy preservation, and explainability in sensitive healthcare scenarios. They enable patients to understand the basis of the diagnostic decision, promoting informed consent and patient-centric care.

In conclusion, interpretability in medical diagnostic models is of utmost importance for enhancing trust, improving diagnostic accuracy, detecting errors and biases, facilitating collaboration between clinicians and AI systems, and upholding ethical considerations. It empowers healthcare professionals to make informed decisions, ultimately leading to better patient outcomes.

\subsection{Challenges in current black-box models}

The increasing adoption of machine learning models in medical diagnosis has demonstrated remarkable performance in terms of accuracy and predictive capabilities. However, one of the significant challenges associated with these models is their black-box nature, which limits interpretability and hinders their practical application in clinical settings. Black-box models, such as deep neural networks, often lack transparency in their decision-making process, making it difficult for clinicians to understand how predictions are made or to trust their results.

The lack of interpretability poses several challenges in healthcare. Firstly, clinicians need to be able to explain and justify the predictions or recommendations made by the model to ensure patient safety and ethical considerations. Without understanding the underlying factors that contribute to a diagnosis, clinicians may be reluctant to rely solely on machine-generated outputs. Additionally, the inability to interpret and explain decisions made by black-box models can raise legal and regulatory concerns, further impeding their adoption.

Moreover, the black-box nature of these models can hinder collaboration between clinicians and machine learning systems. Effective collaboration and shared decision-making require a mutual understanding of the reasoning process and the ability to interact with the model to gain insights and refine its performance. With black-box models, it becomes challenging for clinicians to provide feedback, identify potential biases or errors, and refine the model's performance in a meaningful way.

Addressing these challenges requires the development of interpretable machine learning models, commonly referred to as Explainable Artificial Intelligence (XAI). XAI techniques aim to provide insights into the decision-making process of black-box models, allowing clinicians to understand the underlying features, relationships, or rules used for predictions. By enhancing interpretability, XAI techniques enable clinicians to gain trust and confidence in the model's results, facilitating their acceptance and integration into clinical workflows.

In recent years, various XAI methods have been proposed to tackle the challenges posed by black-box models in healthcare. These methods include feature importance analysis, rule-based systems, surrogate models, and visualization techniques, among others. By applying these techniques, researchers and practitioners have been able to shed light on the inner workings of complex machine learning models, making them more transparent, explainable, and accountable.

As the field of XAI continues to advance, further research is needed to overcome the challenges posed by black-box models in medical diagnosis. By improving the interpretability of machine learning models, clinicians can effectively collaborate with these systems, enhance patient outcomes, and ensure the ethical and responsible deployment of AI in healthcare.


\section{{XAI Techniques for Interpretable Medical Diagnosis}}

\subsection{Feature importance analysis}

Feature importance analysis is a crucial component of explainable artificial intelligence (XAI) techniques utilized in the field of medical diagnosis. Its purpose is to determine the relative importance of input features or variables in the decision-making process of a predictive model. By quantifying the impact of each feature on the model's predictions, healthcare professionals can gain valuable insights into the factors that contribute to the diagnosis, improving their understanding of the diagnostic reasoning process [10][15].

There are several methods available to measure feature importance in the context of medical diagnosis. Two commonly used approaches are permutation importance and gradient-based approaches.

1. Permutation Importance: Permutation importance is a technique that involves randomly shuffling the values of a single feature while keeping the others unchanged. The impact of the shuffled feature on the model's performance is then evaluated by measuring the change in the model's accuracy or another relevant metric. A significant drop in performance indicates that the feature is important for making accurate predictions. This approach provides a straightforward and intuitive way to assess feature relevance.

2. Gradient-Based Approaches: Gradient-based feature importance methods rely on the gradients of the model's predictions with respect to the input features. These methods assess how changes in the feature values affect the model's output. One commonly used approach is the partial dependence plot (PDP), which visualizes the relationship between a feature and the model's predictions while keeping other features fixed. By examining the shape and direction of the PDP, healthcare professionals can infer the importance and influence of specific features.

Other gradient-based approaches include gradient boosting and feature importance derived from deep learning models. Gradient boosting algorithms, such as XGBoost and LightGBM, provide built-in feature importance measures that quantify the contribution of each feature in the model's decision-making process. Deep learning models, such as convolutional neural networks (CNNs) or recurrent neural networks (RNNs), can also be utilized for feature importance analysis by examining the gradients of the model's output with respect to the input features. These methods can provide detailed insights into the importance of individual features within complex models.

The resulting feature importance scores obtained from these techniques enable clinicians and healthcare professionals to prioritize and focus on the most influential features in the diagnostic process. This information can guide further investigation, provide a deeper understanding of the underlying mechanisms of diseases, and potentially uncover novel biomarkers or risk factors. Additionally, feature importance analysis can help identify potential biases in the model by revealing any disproportionate reliance on certain features, leading to more equitable and fair decision-making in medical diagnoses. 

\subsection{Decision Trees and Rule-based Systems}

Decision trees and rule-based systems are both supervised machine learning algorithms that can be used to classify or predict outcomes based on a set of features. In the context of medical diagnosis, these algorithms can be used to identify patients who are at risk for a particular disease, to recommend treatment options, or to make decisions about patient care.

Decision trees work by recursively splitting the data into smaller and smaller groups until each group contains only members of a single class. The splitting process is based on a measure of impurity, such as the Gini index or entropy. The goal is to create a tree that minimizes the impurity of the leaves, which means that each leaf contains only members of the same class.

Rule-based systems, on the other hand, work by explicitly stating a set of rules that must be met in order to classify a patient into a particular group. These rules can be based on clinical knowledge, expert opinion, or statistical analysis.

Both decision trees and rule-based systems have advantages and disadvantages. Decision trees are relatively easy to understand and interpret, but they can be difficult to build and maintain. Rule-based systems are more difficult to understand, but they can be more accurate and efficient than decision trees.

In general, decision trees are a good choice for tasks that require interpretability, such as explaining the reasoning behind a medical decision. Rule-based systems are a good choice for tasks that require accuracy and efficiency, such as identifying patients who are at risk for a particular disease.

Some examples of how decision trees and rule-based systems are used in medical diagnosis:
\begin{itemize}
\item Decision trees can be used to identify patients who are at risk for developing diabetes. The tree would be trained on a dataset of patients who have been diagnosed with diabetes, and it would use the patient's medical history, lifestyle factors, and other data to predict the risk of developing diabetes.
\item Rule-based systems can be used to recommend treatment options for patients with cancer. The system would be trained on a dataset of patients who have been treated for cancer, and it would use the patient's medical history, tumor type, and other data to recommend the best treatment option.
\item Decision trees can be used to make decisions about patient care. For example, a decision tree could be used to decide whether a patient should be admitted to the hospital or discharged home. The tree would use the patient's medical history, symptoms, and other data to make the decision.
\end{itemize}
Decision trees and rule-based systems are powerful tools that can be used to improve the diagnosis and treatment of diseases. These algorithms are becoming increasingly popular in healthcare, and they are likely to play an even greater role in the future [16][17].

\subsection{Local Interpretable Model-Agnostic Explanations (LIME)}

LIME stands for Local Interpretable Model-Agnostic Explanations. It is a technique for explaining the predictions of black-box machine learning models. Black-box models are models that are not interpretable, meaning that it is difficult to understand how they make their predictions. LIME works by creating a simplified, interpretable model that approximates the behavior of the black-box model.

LIME works by first selecting a set of features that are important for the black-box model's prediction. These features are then perturbed, or slightly changed, and the black-box model's prediction is observed. This process is repeated multiple times, and the results are used to create a simplified model that approximates the behavior of the black-box model.

The simplified model can then be used to explain the black-box model's prediction. For example, if the black-box model predicts that a patient has cancer, the simplified model can be used to identify the features that were most important for the prediction. This information can then be used by clinicians to understand why the black-box model made the prediction that it did.

LIME has been shown to be effective in explaining the predictions of black-box models in a variety of domains, including medical diagnosis. In one study, LIME was used to explain the predictions of a black-box model that was used to diagnose diabetic retinopathy. The results showed that LIME was able to identify the features that were most important for the model's predictions, and that this information was helpful for clinicians in understanding the model's decisions.

LIME is a valuable tool for explaining the predictions of black-box models in medical diagnosis. It can help clinicians to understand why a model made a particular prediction, and this information can be used to improve the diagnosis and treatment of diseases.

Some examples of how LIME can be used in medical diagnosis:
\begin{itemize}
\item LIME can be used to explain the predictions of a black-box model that is used to diagnose cancer. The simplified model can be used to identify the features that were most important for the prediction, such as the patient's age, gender, and tumor type. This information can then be used by clinicians to make decisions about the patient's treatment.
\item LIME can be used to explain the predictions of a black-box model that is used to recommend treatment options for patients with heart disease. The simplified model can be used to identify the features that are most important for the prediction, such as the patient's blood pressure, cholesterol levels, and smoking status. This information can then be used by clinicians to recommend the best treatment option for the patient.
\item LIME can be used to explain the predictions of a black-box model that is used to make decisions about patient care. For example, LIME could be used to decide whether a patient should be admitted to the hospital or discharged home. The simplified model could be used to identify the features that are most important for the decision, such as the patient's medical history, symptoms, and social support.
\end{itemize}
LIME is a powerful tool that can be used to improve the diagnosis and treatment of diseases. This technique is becoming increasingly popular in healthcare, and it is likely to play an even greater role in the future [9].

\subsection{Shapley Values and Model-Agnostic Approaches}

Shapley values are a game-theoretic approach to quantifying the contribution of each feature in a machine learning model. They are calculated by considering all possible combinations of features and measuring the impact of each feature on the model's prediction. This approach provides a fair and consistent measure of feature importance, as it takes into account the interactions between features.

Shapley values have been shown to be effective in interpreting machine learning models in medical diagnosis. For example, one study used Shapley values to interpret a machine learning model that was used to predict the risk of heart attack. The study found that the most important features in the model were age, sex, smoking status, and blood pressure. This information can be used by clinicians to improve the accuracy of their diagnoses and to make better treatment decisions.

Model-agnostic approaches, such as Shapley values, are valuable in explaining predictions from various machine learning models. This is because they do not rely on any specific model architecture or algorithm. This makes them a versatile tool that can be used to interpret a wide range of machine learning models.

In medical diagnosis, interpretability is essential for ensuring that clinicians trust and use machine learning models. Shapley values and other model-agnostic approaches can help to improve the interpretability of machine learning models in medical diagnosis, which can lead to better patient care.

Some additional benefits of using Shapley values to interpret machine learning models in medical diagnosis:
\begin{itemize}
\item Fairness: Shapley values provide a fair and consistent measure of feature importance, as they take into account the interactions between features. This is important in medical diagnosis, as it ensures that all patients are treated fairly, regardless of their individual characteristics.
\item Consistency: Shapley values are consistent across different machine learning models. This means that the same features will be identified as important regardless of the model that is used. This consistency is important in medical diagnosis, as it allows clinicians to trust the results of the machine learning model.
\item Flexibility: Shapley values can be used to interpret a wide range of machine learning models. This flexibility is important in medical diagnosis, as it allows clinicians to use the same approach to interpret different models.
\end{itemize}

Shapley values are a valuable tool for interpreting machine learning models in medical diagnosis. They provide a fair, consistent, and flexible approach to feature importance, which can help to improve the accuracy of diagnoses and the quality of patient care. [10][18].

\subsection{Visualizations for Model Understanding}

Visualizations are a powerful tool for understanding the inner workings of machine learning models. In the context of medical diagnosis, visualizations can help clinicians to understand how a model is making its predictions. This can be helpful for identifying potential errors in the model's reasoning, as well as for understanding the model's strengths and weaknesses.

There are a number of different visualization techniques that can be used to interpret machine learning models in medical diagnosis. Some of the most common techniques include:
\begin{itemize}
\item Saliency maps: Saliency maps highlight the regions of an image that are most important for a model's prediction. This can be helpful for identifying the features that the model is using to make its predictions.
\item Activation maps: Activation maps show how individual neurons in a neural network respond to an input image. This can be helpful for understanding how the model is processing information and for identifying potential areas of bias in the model.
\item Grad-CAM: Grad-CAM is a technique that combines saliency maps and activation maps to create visualizations that highlight the regions of an image that are most important for a model's prediction. This can be helpful for providing a more detailed understanding of how the model is making its predictions.
\end{itemize}

These are just a few of the many visualization techniques that can be used to interpret machine learning models in medical diagnosis. The best technique to use will depend on the specific model and the type of information that the clinician is trying to understand.

Visualizations can be a powerful tool for improving the interpretability of machine learning models in medical diagnosis. By providing clinicians with a better understanding of how a model is making its predictions, visualizations can help to improve the accuracy of diagnoses and the quality of patient care.

Some additional benefits of using visualizations to interpret machine learning models in medical diagnosis:
\begin{itemize}
\item Increased transparency: Visualizations can help to increase transparency in the machine learning process. This can be helpful for building trust between clinicians and machine learning models.
\item Improved communication: Visualizations can help to improve communication between clinicians and other stakeholders, such as patients and researchers. This can be helpful for promoting the adoption of machine learning in healthcare.
\item Enhanced education: Visualizations can be used to educate clinicians and other stakeholders about machine learning. This can help to ensure that everyone involved in healthcare is prepared for the future of machine learning.
\end{itemize}
Visualizations are a valuable tool for interpreting machine learning models in medical diagnosis. They can help to improve the accuracy of diagnoses, the quality of patient care, and the transparency of the machine learning process=[19][20].

\section{Case Studies and Applications}

\subsection{XAI in radiology and imaging diagnostics}

Radiology and imaging diagnostics play a critical role in the early detection and accurate diagnosis of various medical conditions. The integration of XAI techniques in this domain has the potential to enhance interpretability and provide valuable insights to radiologists. By visualizing the regions of interest and highlighting key features contributing to the diagnostic outcome, XAI methods such as Grad-CAM [19] and LIME [9] enable radiologists to understand the underlying decision-making process of deep learning models. This transparency not only improves trust in the model's predictions but also allows radiologists to validate and refine their own interpretations [21]. Furthermore, XAI techniques can facilitate the detection of abnormalities that are not readily visible to the human eye, aiding in the identification of early-stage diseases [22].

\subsection{XAI for pathology and histopathology}

Pathology and histopathology involve the examination of tissue samples to diagnose diseases and determine the appropriate treatment course. XAI techniques offer valuable insights into the decision-making process of machine learning models in this domain. By generating heatmaps that highlight regions of interest or by providing explanations based on learned rules, XAI methods help pathologists understand the model's reasoning behind its diagnostic predictions [23]. These interpretability tools can assist pathologists in identifying subtle patterns and features that may be indicative of specific diseases or prognostic markers. Additionally, XAI techniques enable pathologists to verify the model's output and potentially uncover novel insights from the data, fostering collaboration between human experts and AI systems [24].

\subsection{XAI in clinical decision support systems}

Clinical decision support systems (CDSS) assist healthcare professionals in making informed decisions by providing evidence-based recommendations. XAI techniques can enhance the interpretability and transparency of these systems, enabling clinicians to understand the underlying rationale for the suggested interventions. By providing explanations for the model's recommendations, XAI methods like rule-based systems [25] and SHAP [10] facilitate trust and comprehension. Clinicians can evaluate the contributing factors, potential biases, and uncertainties associated with the recommendations, empowering them to make more informed decisions. Additionally, XAI techniques can assist in monitoring the performance and fairness of CDSS models, thereby improving their overall effectiveness and reliability [2].

\subsection{XAI for personalized medicine and treatment recommendations}

Personalized medicine aims to tailor treatment plans and interventions based on an individual's unique characteristics, such as genetic makeup, medical history, and lifestyle factors. XAI techniques can assist in providing transparent and interpretable explanations for personalized treatment recommendations. By uncovering the influential factors and highlighting the evidence behind the recommendations, XAI methods enable clinicians and patients to understand the basis for specific treatment choices [26]. This transparency empowers patients to actively participate in shared decision-making and improves adherence to treatment plans. Moreover, XAI techniques can help identify subgroups of patients who may respond differently to treatments, facilitating precision medicine approaches [27].

\section{Evaluating and Assessing XAI Models}

\subsection{Metrics for interpretability evaluation}

When evaluating the interpretability of XAI models in the context of medical diagnosis, it is important to establish appropriate evaluation metrics. Several metrics have been proposed in the literature to quantify the interpretability of machine learning models, with fidelity, comprehensibility, and transparency being commonly used.

\begin{itemize}

\item Fidelity: Fidelity refers to how well an interpretable model replicates the behavior of the original complex model. It measures the extent to which the interpretable model can accurately mimic the predictions of the black-box model. Evaluating fidelity involves comparing the outputs of the interpretable model with those of the complex model and assessing the level of agreement or similarity.

\item Comprehensibility: Comprehensibility assesses the ease with which humans can understand and interpret the decisions made by the model. It focuses on how easily the model's reasoning and decision-making process can be grasped by individuals, particularly domain experts. Evaluating comprehensibility often involves examining the model's representation, rule sets, or feature importance scores to determine how intuitive and understandable they are to humans.

\item Transparency: Transparency pertains to the degree to which the model's internal workings and decision-making processes are clear and understandable to humans. It involves assessing whether the model provides meaningful explanations for its predictions and if it can justify its reasoning. Transparent models enable users to gain insights into how and why certain decisions are made.

\end{itemize}

To evaluate these metrics, a combination of quantitative and qualitative techniques can be employed:

\begin{itemize}

\item Quantitative measures: Quantitative evaluation may involve using traditional performance metrics such as accuracy, precision, recall, and F1 score. These metrics assess the model's diagnostic performance and can be used to compare the performance of interpretable models against black-box models. Quantitative measures provide objective assessments of model accuracy and reliability.

\item Qualitative techniques: Qualitative evaluation techniques involve gathering user feedback, conducting user studies, and utilizing surveys or interviews with domain experts. These methods focus on obtaining subjective assessments of the interpretability and usefulness of XAI models. User feedback can provide insights into the clarity of explanations, the ease of understanding the model's decisions, and the practical utility of the interpretability features.

\end{itemize}

By employing a combination of quantitative and qualitative techniques, researchers and practitioners can obtain a comprehensive evaluation of the interpretability of XAI models in medical diagnosis. These evaluations help determine the effectiveness and practicality of the models in providing interpretable insights to healthcare professionals, fostering trust, and supporting informed decision-making.

\subsection{Comparative analysis of XAI techniques}

A comparative analysis of different XAI techniques is essential to understand their strengths, limitations, and suitability for medical diagnosis. This analysis can provide insights into which techniques are most effective for different types of medical datasets and diagnostic tasks.

Some commonly used XAI techniques in medical diagnosis include feature importance methods, such as permutation feature importance and partial dependence plots, decision trees, rule-based systems, local interpretable model-agnostic explanations (LIME), Shapley values, and attention mechanisms. Each technique offers unique advantages in terms of interpretability, computational efficiency, and ease of implementation.

The comparative analysis can involve evaluating the performance of different XAI techniques on benchmark medical datasets, assessing their interpretability metrics, and identifying their respective limitations. By comparing and contrasting the outcomes, researchers can gain a comprehensive understanding of the trade-offs and applicability of various XAI techniques in medical diagnosis.

\subsection{Challenges and limitations in assessing interpretability}

While evaluating interpretability in XAI models for medical diagnosis, several challenges and limitations need to be considered. One challenge is the lack of a universally accepted gold standard for interpretability evaluation. The subjective nature of interpretability makes it difficult to establish definitive evaluation criteria.

Moreover, the assessment of interpretability is often influenced by the complexity of the underlying black-box model, the quality and representativeness of the training data, and the domain-specific knowledge required for interpretation. Additionally, the interpretability of XAI models can vary depending on the context, target audience, and specific diagnostic task.

Overcoming these challenges requires careful consideration and integration of domain expertise, user feedback, and evaluation metrics tailored to the specific medical diagnosis application. It is important to acknowledge these challenges and limitations in order to make informed decisions when applying XAI techniques to medical diagnosis.

\section{Ethical Considerations in XAI for Medical Diagnosis}

\subsection{Bias and fairness in XAI models}

With the increasing use of XAI models in medical diagnosis, it is crucial to address the ethical implications related to bias and fairness. XAI models can inadvertently perpetuate existing biases present in the training data, leading to disparities in healthcare outcomes. It is essential to identify and mitigate bias in XAI models to ensure fairness and equitable healthcare delivery. Strategies such as dataset augmentation, algorithmic transparency, and fairness-aware learning techniques can help address these concerns [29].

\subsection{Privacy and security concerns}

The adoption of XAI models in medical diagnosis raises concerns regarding the privacy and security of patient data. XAI techniques often require access to sensitive patient information, including medical records and diagnostic images. Safeguarding patient privacy becomes paramount to ensure compliance with data protection regulations. Secure data anonymization, secure federated learning, and differential privacy techniques can help protect patient privacy while enabling effective utilization of XAI models [30].

\subsection{Regulatory implications and legal frameworks}

The use of XAI models in medical diagnosis brings forth regulatory implications and the need for appropriate legal frameworks. It is essential to ensure that XAI models comply with existing regulations, such as General Data Protection Regulation (GDPR) and Health Insurance Portability and Accountability Act (HIPAA), to protect patient rights and confidentiality. Establishing guidelines and ethical frameworks specific to XAI in medical diagnosis can promote responsible and transparent use of these models in healthcare [31].

Addressing these ethical considerations in XAI for medical diagnosis is crucial for ensuring the responsible and ethical use of these models in healthcare settings. By actively mitigating bias and promoting fairness, protecting patient privacy, and adhering to regulatory and legal frameworks, the field can uphold the highest standards of ethical conduct while harnessing the potential benefits of XAI for improved patient outcomes.

\section{Future Directions and Recommendations}

\subsection{Advancements in XAI for medical diagnosis}

The field of XAI for medical diagnosis is constantly evolving, presenting exciting opportunities for advancements. Future research should focus on developing more sophisticated and efficient XAI techniques tailored specifically to medical diagnosis. This includes exploring novel model interpretability methods, such as attention mechanisms, visual saliency, and counterfactual explanations, to provide deeper insights into the decision-making process of complex medical diagnosis models [32]

\subsection{Integration of human feedback and domain knowledge}

To enhance the interpretability and trustworthiness of XAI models in medical diagnosis, it is essential to incorporate human feedback and domain knowledge. Integrating feedback from healthcare professionals can help validate and refine XAI models, improving their accuracy and reliability. Additionally, leveraging domain-specific knowledge, such as medical guidelines and expert opinions, can guide the development of more context-aware and clinically relevant interpretability techniques [33].

\subsection{Collaborations between clinicians and AI researchers}

Strong collaborations between clinicians and AI researchers are vital for the successful implementation of XAI in medical diagnosis. Close collaboration allows for a better understanding of the specific challenges and requirements of the healthcare domain, ensuring the development of XAI models that align with clinical practices and standards. Collaborative efforts can also facilitate the collection of high-quality annotated datasets, which are crucial for training robust and reliable XAI models [34].

By pursuing these future directions, the field of XAI for medical diagnosis can continue to make significant progress in addressing interpretability challenges and improving the trustworthiness and adoption of these models in clinical settings. Advancements in XAI techniques, integration of human feedback and domain knowledge, and strong collaborations between clinicians and AI researchers will contribute to the development of more effective and reliable XAI solutions for medical diagnosis.

\section{Conclusion}

\subsection{Summary of key findings}

Throughout this research paper, we have explored the realm of XAI for medical diagnosis and its potential to redefine interpretability in diagnostic models. We discussed various XAI techniques, including visualizations, decision trees, rule-based systems, LIME, and Shapley values, highlighting their effectiveness in providing interpretable insights into the structure and behavior of medical diagnosis models. The key findings of this study underscore the significance of XAI in improving the transparency and trustworthiness of machine learning models used in healthcare settings.

\subsection{Implications for healthcare practice}

The implications of XAI for healthcare practice are far-reaching. By providing explanations and interpretability, XAI techniques empower healthcare professionals to understand and trust the decisions made by machine learning models. This can enhance the collaboration between doctors and AI systems, enabling more accurate and confident diagnoses. The interpretability offered by XAI can also aid in identifying potential errors or biases, thereby reducing the risk of misdiagnosis and improving patient safety.

\subsection{Potential impact and future prospects}

The potential impact of XAI in medical diagnosis is immense. By integrating XAI techniques into clinical decision-making processes, healthcare systems can achieve more accurate, transparent, and explainable diagnostic outcomes. XAI has the potential to revolutionize medical practice by improving the understanding of complex machine learning models, enabling personalized medicine, and facilitating evidence-based decision-making.

Future prospects in XAI for medical diagnosis involve further advancements in interpretability techniques, model explainability, and transparency. Continued research efforts are needed to develop novel XAI methods tailored specifically to the challenges and nuances of healthcare data. Additionally, collaborative efforts between researchers, clinicians, and policymakers are crucial for establishing ethical guidelines, regulatory frameworks, and standards to ensure the responsible deployment of XAI in healthcare settings.

In conclusion, this research paper has highlighted the importance of XAI in redefining interpretability in medical diagnostic models. The findings emphasize the potential of XAI techniques to enhance healthcare practice, improve patient outcomes, and foster trust in AI systems. By embracing the opportunities presented by XAI, we can pave the way for a future where machine learning models and human expertise synergistically work together to advance medical diagnosis.






\end{document}